# Evaluating Hospital Case Cost Prediction Models Using Azure Machine Learning Studio

Alexei Botchkarev

## Abstract

Ability for accurate hospital case cost modelling and prediction is critical for efficient health care financial management and budgetary planning. A variety of regression machine learning algorithms are known to be effective for health care cost predictions. The purpose of this experiment was to build an Azure Machine Learning Studio tool for rapid assessment of multiple types of regression models. The tool offers environment for comparing 14 types of regression models in a unified experiment: linear regression, Bayesian linear regression, decision forest regression, boosted decision tree regression, neural network regression, Poisson regression, Gaussian processes for regression, gradient boosted machine, nonlinear least squares regression, projection pursuit regression, random forest regression, robust regression, robust regression with mm-type estimators, support vector regression. The tool presents assessment results in a single table using 22 performance metrics: CoD, GMRAE, MAE, MAPE, MASE, MdAE, MdAPE, MdRAE, ME, MPE, MRAE, MSE, NRMSE_mm, NRMSE_sd, RAE, RMdSPE, RMSE, RMSPE, RSE, sMAPE, SMdAPE, SSE.

Evaluation of regression machine learning models for performing hospital case cost prediction demonstrated advantage of robust regression model, boosted decision tree regression and decision forest regression.

The operational tool has been published to the web and openly available at Azure MLS for experiments and extensions.

---------------------------------------------------------

**Note: This the 2nd version of the working paper.**

It is based on an updated Azure MLS operational tool:

> Botchkarev, A. (2018). Revision 2 Integrated tool for rapid assessment of multi-type regression machine learning models. Experiment in Microsoft Azure Machine Learning Studio. Azure AI Gallery. *https://gallery.azure.ai/Experiment/Revision-2-Integrated-tool-for-rapid-assessment-of-multi-type-regression-machine-learning-models*

In the 2nd Revision of the tool, all regression models were assessed with a newly developed Enhanced Evaluation Model (EEM) module. Number of evaluation performance metrics has been increased to 22.

EEM is available as Azure MLS experiment:

> Botchkarev, A. (2018). Enhanced model evaluation with multiple performance metrics for regression analysis. Experiment in Microsoft Azure Machine Learning Studio. Azure AI Gallery. *https://gallery.azure.ai/Experiment/Enhanced-model-evaluation-with-multiple-performance-metrics-for-regression-analysis*

The details of the EEM are presented in:

> Botchkarev, A. (2018). Evaluating Performance of Regression Machine Learning Models Using Multiple Error Metrics in Azure Machine Learning Studio (May 12, 2018). Available at SSRN: *http://ssrn.com/abstract=3177507*

Also, noted errors of the earlier version have been fixed.

## Introduction

Information on patient-level cost of hospital treatment is vital for cost-effectiveness analysis (e.g. Alpenberg & Scarbrough, 2015; Teague, et al, 2011). Ability for accurate hospital case cost modelling and prediction is important for efficient health care financial management, budgetary planning and analysis purposes (e.g. Osnabrugge, et al, 2014; Corral, et al, 2016).

Various types of regression machine learning algorithms have been examined for health care cost predictions. For example, Botchkarev & Andru (2008) investigated multiple linear regression, Sushmita, et al (2015) studied regression tree, M5 model tree and random forest, Srinivasan, Currim & Ram, (2017) conducted experiments with hierarchical decision trees, random forest, linear regression and gradient-boosted trees.

Commonly, research teams investigate low number of regression types in one experiment: two to four. Also, specifics of algorithms implementation, use of different performance metrics and particularities of employed data sets make results of the studies difficult to compare, interpret and reproduce.

To overcome this problem, current research focused on developing a tool that would embrace multiple types of regression models in the same experiment, use diverse (but the same for all algorithms) performance metrics, allow for easy change of input data sets.

## Methodology

### Objective

The objective of the experiment was to build an Azure MLS tool and compare various types of regression machine learning models for hospital case cost prediction and select models with higher performance for further examination.

### Experiment Platform

Microsoft Azure Machine Learning Studio (https://studio.azureml.net) has been selected as a platform for building a tool and conducting experiments. Selection of the Azure MLS was motivated by the following features directly contributing to the objectives of the study:

- Cloud-based machine learning as a service.
- Web-based solution – user needs browser only to work with the system. No set up, installation and maintenance concerns or complications.
- Easy to use drag and drop canvas interface for intuitively clear aggregating computing modules into an experiment.
- Several ready to use built-in regression modules.
- Flexibility of using R and Python languages to code experiments.
- Ability to integrate functions from R packages.
- Easy to access, password-protected integrated development environment.
- Capability to publish results of experiments to the web.
- Capability to re-use published experiments or their components.
- Low fee (pay as you go) or even free service.

**Regression models included in the experiment**

Azure built-in models:

- linear regression,
- Bayesian linear regression,
- decision forest regression,
- boosted decision tree regression,
- neural network regression and
- Poisson regression.

Azure MLS built-in algorithms were complemented by models developed using R language modules: Execute R Script and Create R model:

- Gaussian processes for regression,
- gradient boosted machine,
- nonlinear least squares regression,
- projection pursuit regression,
- random forest regression,
- robust regression,
- robust regression with mm-type estimators,
- support vector regression (support vector machine).

**Data analysis**

The experiment used a simulated data set intended to mimic hospital information. The data set had the following columns (features): Intervention, Diagnosis, Case Mix Group (CMG), Gender, Age Group (Age Gr), Relative Intensity Weight (RIW), Length of Stay (LOS), Cost. Total number of rows in the data set is 7,000.

An overview of the data set characteristics is provided in Appendix 1. Note that the data set is **simulated** and no warranties are provided as to the validity of the data and how closely it simulates real-world information.

**Out of scope.** Building operational HCC model and making actual HCC predictions was out of scope.

**Experiment phases**

**Phase 1.** The purpose of the experiment Phase 1 was to investigate predictive power of the features (columns) of the data set. Azure MLS Filter-Based Feature Selection module was used to score all features with several criteria and select most important. This phase allowed identifying which columns must be used in the experiment because they contribute the most to the model results and which should not be used.

**Phase 2.** The purpose of the experiment Phase 2 was to compare Azure MLS built-in types of regression machine learning models: linear regression, Bayesian linear regression, decision forest regression, boosted decision tree regression, neural network regression and Poisson regression.

In Phase 2, all models were used with Azure default parameters. For convenience, outputs of all models are presented in a single table. In the modules where custom random seed number is allowed, it was set to 12345.

Output provides a single combined table of the results for all regression types models used in the experiment and the names of algorithms. R script was used to build a combined table of results. Note that outputs of the Evaluate modules for different algorithms have different format. Hence, Execute R script modules have different scripts.

**Phase 3.** Fine tune models' parameters. The purpose was to tune regression model parameters to enhance performance metrics. In some sources, this process is called model optimization. Azure Tune Model Hyperparameters module was used for perform optimization.

**Phase 4.** Assessing models' tolerance to change. The purpose of Phase 4 was to examine the reaction of the models on the changes introduced into computing environment and datasets, e.g. setting different seed numbers in model calculations, changing percentage of data split between the data sets used for testing and scoring, changing (adding or removing) features of the data set.

**Phase 5.** The objective of Phase 5 was to experiment with all 14 regression types. Both available in Azure as built-in modules and developed with R language. Azure MLS provides a "Create R Model" module to develop algorithms using R language and capitalize on a vast variety of regression methods implemented in R Studio. The following types of regression were implemented: Gaussian processes for regression, gradient boosted machine (GMB), nonlinear least squares regression, projection pursuit regression, random forest regression, robust regression, robust regression with mm-type estimators, support vector regression (support vector machine). Table 1 shows which R packages and functions were used in the experiment.

**Table 1. R packages and functions were used in the experiment**

| Regression Type | R Package | Function |
|---|---|---|
| Gaussian Processes for Regression | kernlab | gausspr |
| Gradient Boosted Machine (GBM) | caret | train (gbm) |
| Nonlinear Least Squares Regression | stats | nls |
| Projection Pursuit Regression | stats | ppr |
| Random Forest Regression | randomForest | randomForest |
| Robust Regression | MASS | rlm |
| Robust Regression with MM-Type Estimators | robustbase | lmrob |
| Support Vector Regression (Support Vector Machine) | E1071 | svm |

**Performance metrics.** In phases 1 to 4, Azure standard performance metrics were used to evaluate models: mean absolute error (MAE), root mean squared error (RMSE), relative absolute error (RAE), relative squared error (RSE), coefficient of determination (CoD).

In Phase 5, Enhanced Evaluate Model module was used which increased number of performance metrics to 22: Coefficient of Determination (CoD), Geometric Mean Relative Absolute Error (GMRAE), Mean Absolute Error (MAE), Mean Absolute Percentage Error (MAPE), Mean Absolute Scaled Error (MASE), Median Absolute Error (MdAE), Median Absolute Percentage Error (MdAPE), Median Relative Absolute Error (MdRAE), Mean Error (ME), Mean Percentage Error (MPE), Mean Relative Absolute Error (MRAE), Mean Squared Error (MSE), Normalized Root Mean Squared Error normalized to the difference between maximum and minimum actual data (NRMSE_mm), Normalized Root Mean Squared Error normalized to the standard deviation of the actual data (NRMSE_sd), Relative Absolute Error (RAE), Root Median Square Percentage Error (RMdSPE), Root Mean Squared Error (RMSE), Root Mean Square Percentage Error (RMSPE), Relative Squared Error (RSE), Symmetric Mean Absolute Percentage Error (sMAPE ), Symmetric Median Absolute Percentage Error (SMdAPE), Sum of Squared Error (SSE).

**Terminology.** In this paper, terms are used per the Azure MLS conventions, e.g. more commonly used R squared is called a coefficient of determination (CoD).

A list of abbreviations is shown in Appendix 6.

*<u>Outcome of the study.</u>* The operational tool has been published to the web (Botchkarev, 2018). It can be used by researchers and practitioners to reproduce the results of this study, conduct experiments with their own data sets or add more regression models to the framework.

## Experiment Results

### Phase 1 Results

The data set has been tested using Azure MLS Filter-Based Feature Selection module with the following selection criteria: Pearson Correlation, Mutual Information, Kendall Correlation, Spearman Correlation, Chi Squared and Fisher Score. Results are presented in Table 2.

**Table 2. Scoring Data Set Features by Various Criteria**

|  | Cost | RIW | LOS | Age_gr | CMG | Gender | Diagn | Interv |
|---|---|---|---|---|---|---|---|---|
| Pearson Correlation | 1 | 0.860 | 0.764 | 0.072 | 0.015 | 0 | 0 | 0 |
|  | Cost | RIW | LOS | CMG | Diagn | Interv | Age_gr | Gender |
| Mutual Information | 1 | 1.852 | 0.332 | 0.308 | 0.287 | 0.244 | 0.053 | 0.007 |
|  | Cost | RIW | LOS | Age_gr | CMG | Gender | Diagn | Interv |
| Kendall Correlation | 1 | 0.957 | 0.494 | 0.156 | 0.023 | 0 | 0 | 0 |
|  | Cost | RIW | LOS | Age_gr | CMG | Gender | Diagn | Interv |
| Spearman Correlation | 1 | 0.986 | 0.663 | 0.223 | 0.030 | 0 | 0 | 0 |
|  | Cost | RIW | Diagn | Interv | LOS | CMG | Age_gr | Gender |
| Chi Squared | 1 | 50509 | 30675 | 24597 | 5424 | 4997 | 754 | 101 |

|  | Cost | RIW | LOS | CMG | Age_gr | Gender | Diagn | Interv |
|---|---|---|---|---|---|---|---|---|
| Fisher Score | 1 | 156.482 | 7.437 | 0.773 | 0.551 | 0 | 0 | 0 |

Based on the feature selection exercise, the training data set has been transformed into six (6) versions with different numbers of included features. Table 3 shows lists of columns in the dataset versions. Please note sequential removal of less important columns between versions. The table also shows the names of the tests in the further experiments based on the data sets they used.

**Table 3 Data set features used in tests**

|  | **Test_1** | **Test_2** | **Test_3** | **Test_4** | **Test_5** | **Test_6** |
|---|---|---|---|---|---|---|
| **Columns/Features** | Cost | Cost | Cost | Cost | Cost | Cost |
|  | RIW | RIW | RIW | RIW | RIW | RIW |
|  | LOS | LOS | LOS | LOS | LOS | LOS |
|  | Age_gr | Age_gr | Age_gr | Age_gr | Age_gr |  |
|  | Gender | Gender | Gender | Gender |  |  |
|  | CMG | CMG | CMG |  |  |  |
|  | Diagn | Diagn |  |  |  |  |
|  | Interv |  |  |  |  |  |

**Phase 2 Results**

The following types of machine learning regression models available in Azure MLS as built-in modules were tested in the experiment: linear regression, Bayesian linear regression, decision forest regression, boosted decision tree regression, neural network regression and Poisson regression.

Results of the Phase 2 experiment are shown in the Appendix 2 Table A2.1.

Graphical comparisons are shown in Fig. 1 for MAE, Fig. 2 for RMSE, and Fig. 3 for CoD.

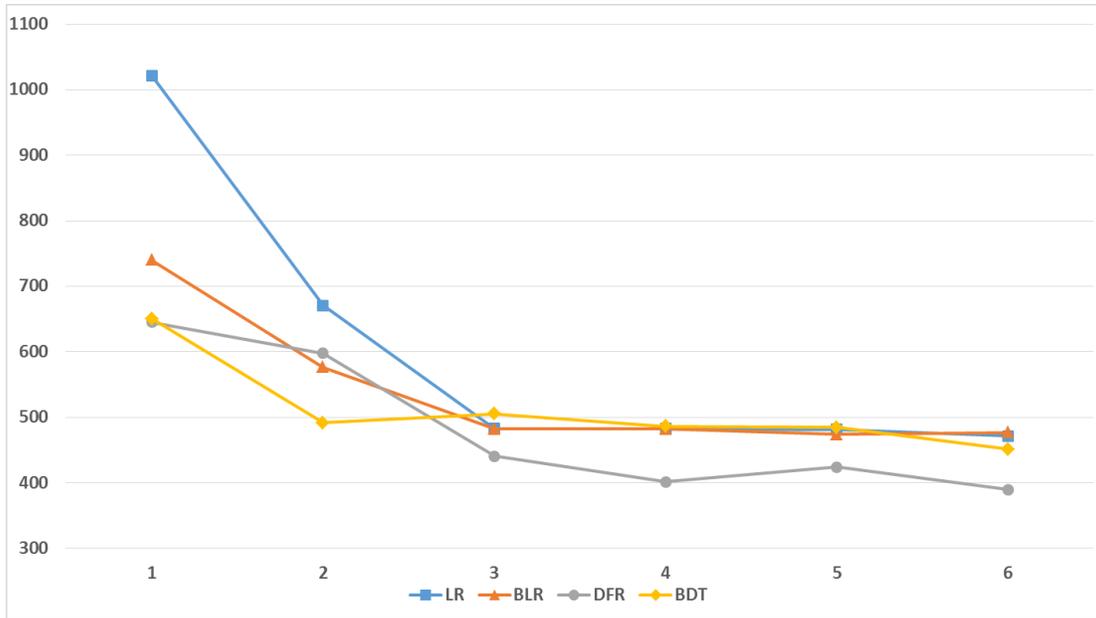

**Fig. 1 Mean absolute error**

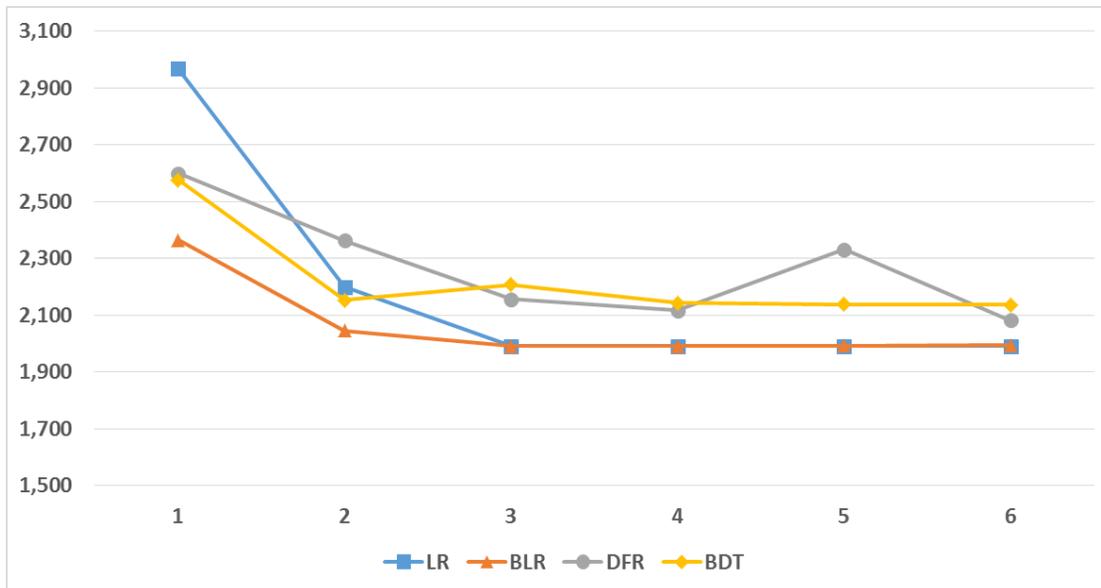

**Fig. 2 Root mean squared error**

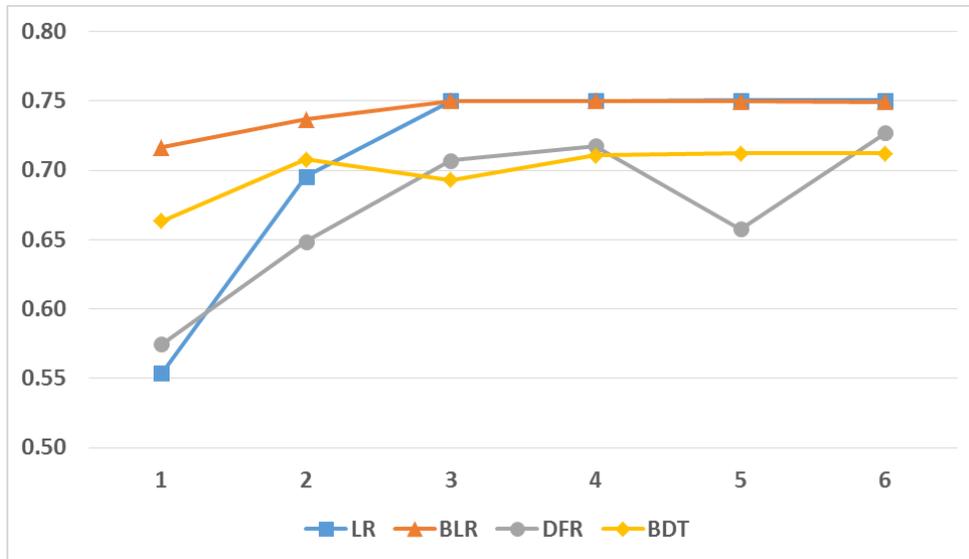

**Fig. 3 Coefficient of determination**

Experiment with Azure built-in modules workflow chart is presented in Fig. 4.

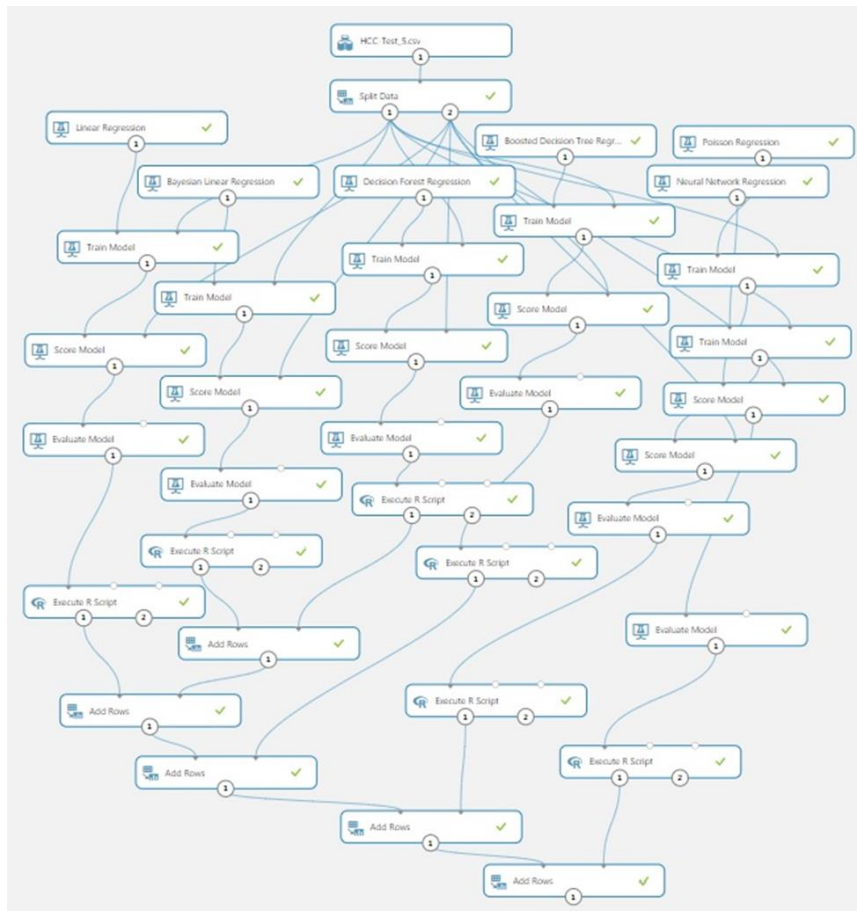

**Fig. 4. Regression Comparison Experiment Flow Chart**

**Results Phase 3**

In Phase 2, all models were trained using Train Model module, specific for each regression type, with default parameters. Azure default parameters may not, and most likely will not, be the best for a specific data set. Tuning model parameters may allow gaining better results in terms of performance metrics criteria. In some sources, this process is called model optimization. Azure Tune Model Hyperparameters module was used for perform optimization.

The structure of the experiment to tune parameters of the Decision Forest Regression is shown in Fig. 5. The chart has two parallel flows. One, for the experiment with default parameters, and the second, on the right side, for tuning the parameters. To run the tuning experiment, certain settings must be made in two modules: Decision Forest Regression (DFR) and Tune Model Hyperparameters (TMH). Settings in the DFR module define the ranges in which parameters will be tuned. There are four parameters: minimum number of samples per leaf node, number of random splits per node, maximum depth of the decision trees and number of decision trees. Settings in the TMH module define the way in which parameter values will be changed in the experiment, e.g. whether the process will involve scanning (sweeping) all possible combinations of points – Entire Grid, or sweeping will be done across random parameter values only – Random Sweeping. Also, optimization criterion must be selected in the TMH module. Mean Absolute Error was used as a criterion in all optimization experiments.

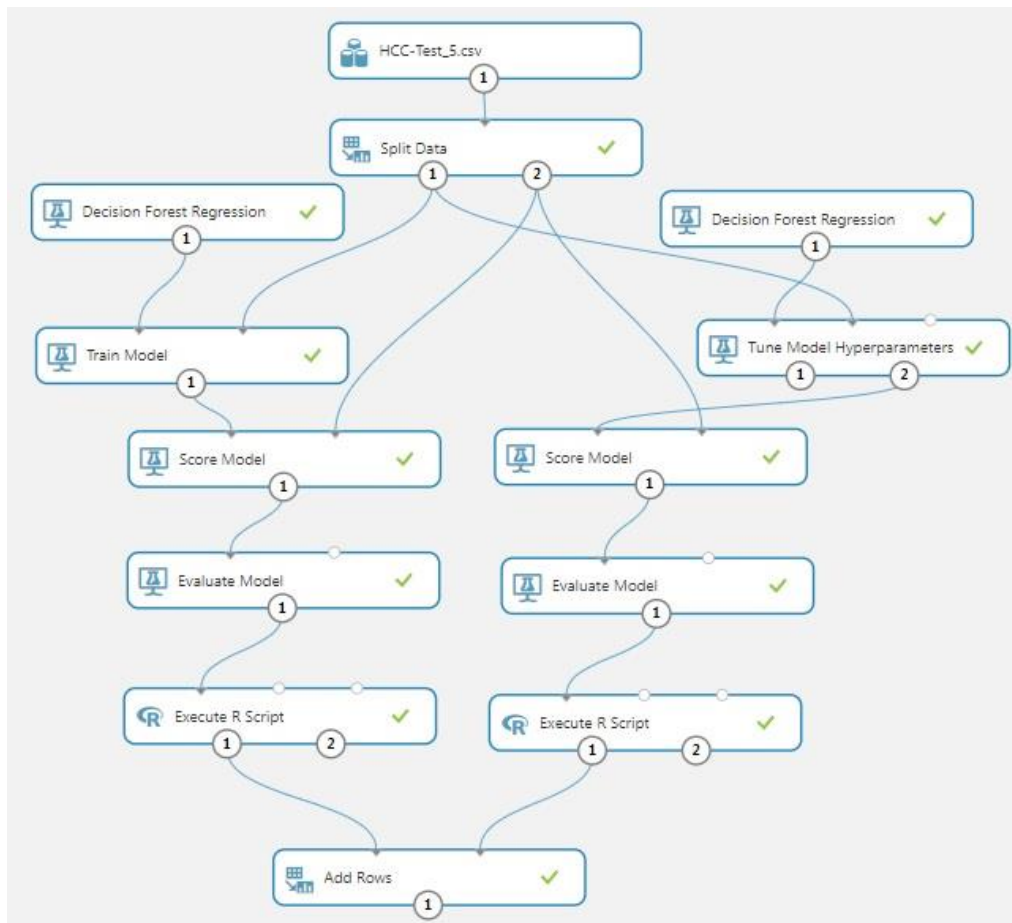

**Fig. 5. Direct Forest Regression Tuning workflow chart**

**Table 4 Decision Forest Regression Tuning Results**

|  | **Performance Metrics** | | | | | **Model Parameters** | | | |
|---|---|---|---|---|---|---|---|---|---|
| **Algorithm** | Mean Absolute Error | Root Mean Squared Error | Relative Absolute Error | Relative Squared Error | Coefficient of Determination | Minimum number of samples per leaf node | Number of random splits per node | Maximum depth of the decision trees | Number of decision trees |
| Decision Forest Regression, Train | 424 | 2332 | 0.23 | 0.34 | 0.66 | 1 | 128 | 32 | 8 |
| Decision Forest Regression, Tune | 403 | 2004 | 0.22 | 0.25 | 0.75 | 7 | 4340 | 368 | 724 |
| Improvement | 5.0% | 14.0% | 5.0% | 26.1% | 13.6% | - | - | - | - |

**Table 5 Boosted Decision Tree Regression Tuning Results**

|  | **Performance Metrics** | | | | | **Model Parameters** | | | |
|---|---|---|---|---|---|---|---|---|---|
| **Algorithm** | Mean Absolute Error | Root Mean Squared Error | Relative Absolute Error | Relative Squared Error | Coefficient of Determination | Number of leaves | Minimum leaf instances | Learning rate | Number of trees |
| Boosted Decision Tree Regression, Train | 484 | 2138 | 0.26 | 0.29 | 0.71 | 20 | 10 | 0.2 | 100 |
| Boosted Decision Tree Regression, Tune | 402 | 1962 | 0.22 | 0.24 | 0.76 | 7 | 15 | 0.097539 | 38 |
| Improvement | 16.9% | 8.2% | 16.9% | 15.8% | 6.4% | - | - | - | - |

Results of tuning for Decision Forest Regression, Boosted Decision Tree Regression are shown in Tables 4 and 5, respectively. These algorithms demonstrated the best ability for tuning. Each table shows performance metrics before and after tuning and optimum parameters that were identified. Also, the tables present levels of improvement for each performance metric compared to the default parameters.

**Results Phase 4**

The following experiments were conducted in Phase 4:

- Changing seed numbers (from initial 12345 to 98642) in the models (where possible).
- Changing percentage of split between testing data set and scoring data set from 0.5 (50% testing, 50% scoring) to 0.7 (70% testing, 30% scoring).
- Changing number of data set features: from initial cost, LOS, RIW and age group to cost, LOS, RIW.

For each option, Azure built-in regression models (5 types) were run and performance metrics were recorded and compared. Tables with metrics data are provided in the appendices and show absolute and relative variances between options:

- Changing seed numbers – Appendix 3.
- Changing percentage of split between testing data set and scoring data set – Appendix 4.
- Changing number of data set features – Appendix 5.

**Results Phase 5**

Combined results of an integrated tool. Fig. 6 shows combined workflow of Azure built-in and R-based regression models. Table 6 shows combined performance metrics for Azure built-in and R-based regression models. Performance of the Boosted Decision Tree Regression and Decision Forest Regression is shown for the optimized versions which moved there position up the list.

Parameters of the R-based modules were not tuned. Most models from R packages were used with default parameters.

Also, modules for multiple linear regression were built with both **glm** and **lm** functions. The results are not shown as they are the same as were gained for the Azure linear regression built-in model. In general, there may be a reason to use **glm** and/or **lm** functions in Azure experiments if there is a need for directly announcing independent variables in order to add or remove individual features. Functions **glm** and **lm** can be used in Azure without calling R libraries (i.e. {stats} or {cars}.

**Table 6 Combined performance metrics for all models**

| metric_abbrev. | full_name | LR | BLR | DFR | BDTR | PR | NNR | RFR | SVR | GRR | RRmm | RR | GBM | PPR | NLSR |
|---|---|---|---|---|---|---|---|---|---|---|---|---|---|---|---|
| CoD | Coefficient of Determination | 0.7505 | 0.7498 | 0.6576 | 0.712 | 0.1769 | -0.0704 | 0.7527 | 0.6852 | 0.5155 | 0.7187 | 0.7189 | 0.7432 | 0.7551 | 0.7505 |
| GMRAE | Geometric Mean Relative Absolute Error | 0.1347 | 0.12 | 0 | 0.0419 | 0.8113 | 0.6322 | 0.1007 | 0.1132 | 0.0568 | 0.0066 | 0.0067 | 0.115 | 0.0771 | 0.1347 |
| MAE | Mean Absolute Error | 481.748 | 473.4953 | 424.2223 | 484.3054 | 1620.5862 | 1594.8297 | 465.3754 | 421.5048 | 554.6658 | 254.557 | 254.5004 | 472.7597 | 430.2943 | 481.748 |
| MAPE | Mean Absolute Percentage Error | 16.9101 | 15.9663 | 6.4989 | 8.7833 | 100.3032 | 60.1384 | 14.7008 | 16.6987 | 9.4456 | 3.0503 | 3.0518 | 18.8039 | 10.6252 | 16.9101 |
| MASE | Mean Absolute Scaled Error | 0.1914 | 0.1881 | 0.1685 | 0.1924 | 0.6437 | 0.6335 | 0.1848 | 0.1674 | 0.2203 | 0.1011 | 0.1011 | 0.1878 | 0.1709 | 0.1914 |
| MdAE | Median Absolute Error | 149.1442 | 129.6546 | 9.1667 | 33.42 | 1175.0388 | 685.4257 | 114.5885 | 156.3304 | 52.2051 | 7.3101 | 7.335 | 135.6288 | 80.3182 | 149.1443 |
| MdAPE | Median Absolute Percentage Error | 10.9705 | 9.0271 | 0.6847 | 2.8282 | 66.5457 | 53.2716 | 7.4764 | 8.2932 | 3.9094 | 0.4761 | 0.4794 | 7.4804 | 5.8489 | 10.9705 |
| MdRAE | Median Relative Absolute Error | 0.1261 | 0.1174 | 0.0091 | 0.0298 | 0.8564 | 0.4501 | 0.089 | 0.1274 | 0.0425 | 0.0064 | 0.0065 | 0.1385 | 0.0823 | 0.1261 |
| ME | Mean Error | 38.4218 | 16.4036 | -46.7055 | 6.3421 | 64.7614 | 1057.3001 | 6.3473 | 133.1441 | 291.3209 | 233.7708 | 233.4515 | -45.8404 | 24.2121 | 38.4219 |
| MPE | Mean Percentage Error | -5.8607 | -8.8064 | -3.7856 | -3.6018 | -82.826 | -14.8161 | -9.7483 | -11.1124 | -2.6764 | 1.016 | 1.0185 | -13.9066 | -5.5076 | -5.8607 |
| MRAE | Mean Relative Absolute Error | 0.6814 | 0.6902 | 0.3027 | 0.4746 | 1.0139 | 2.0807 | 0.4587 | 0.3384 | 0.3606 | 0.0518 | 0.0518 | 0.3082 | 0.4081 | 0.6814 |
| MSE | Mean Squared Error | 3961519.52 | 3971442.311 | 5436572.093 | 4571819.748 | 13067407.76 | 16993900.55 | 3925499.976 | 4998308.675 | 7692362.075 | 4465682.959 | 4463364.639 | 4076863.627 | 3887478.154 | 3961519.431 |
| NRMSE_max_min | Normalized Root Mean Squared Error (norm. to max - min) | 0.0295 | 0.0295 | 0.0345 | 0.0317 | 0.0535 | 0.061 | 0.0293 | 0.0331 | 0.0411 | 0.0313 | 0.0313 | 0.0299 | 0.0292 | 0.0295 |
| NRMSE_sd | Normalized Root Mean Squared Error (norm. to SD) | 0.4995 | 0.5001 | 0.5851 | 0.5366 | 0.9071 | 1.0345 | 0.4972 | 0.561 | 0.696 | 0.5303 | 0.5301 | 0.5067 | 0.4948 | 0.4995 |
| RAE | Relative Absolute Error | 0.2591 | 0.2547 | 0.2282 | 0.2605 | 0.8717 | 0.8579 | 0.2503 | 0.2267 | 0.2984 | 0.1369 | 0.1369 | 0.2543 | 0.2315 | 0.2591 |
| RMdSPE | Root Median Square Percentage Error | 1.0749 | 0.9045 | 0.0682 | 0.284 | 5.0833 | 4.8086 | 0.7315 | 0.8058 | 0.3886 | 0.0476 | 0.048 | 0.7394 | 0.583 | 1.0749 |
| RMSE | Root Mean Squared Error | 1990.3566 | 1992.8478 | 2331.6458 | 2138.1814 | 3614.887 | 4122.3659 | 1981.2875 | 2235.6898 | 2773.5108 | 2113.2163 | 2112.6677 | 2019.1245 | 1971.6689 | 1990.3566 |
| RMSPE | Root Mean Square Percentage Error | 2.6902 | 2.6738 | 2.6372 | 2.421 | 13.399 | 7.2965 | 2.5658 | 2.7828 | 2.1222 | 1.6774 | 1.677 | 3.1995 | 2.1221 | 2.6902 |
| RSE | Relative Squared Error | 0.2495 | 0.2502 | 0.3424 | 0.288 | 0.8231 | 1.0704 | 0.2473 | 0.3148 | 0.4845 | 0.2813 | 0.2811 | 0.2568 | 0.2449 | 0.2495 |
| SMAPE | Symmetric Mean Absolute Percentage Error | 15.8641 | 14.4157 | 5.3109 | 8.0423 | 61.1432 | 57.3569 | 13.1725 | 14.9463 | 9.7217 | 3.439 | 3.4404 | 15.9767 | 10.1177 | 15.8641 |
| SMdAPE | Symmetric Median Absolute Percentage Error | 10.9136 | 9.1048 | 0.6832 | 2.8373 | 61.1301 | 51.6835 | 7.4389 | 8.2134 | 3.8986 | 0.4763 | 0.4799 | 7.4423 | 5.8373 | 10.9136 |
| SSE | Sum of Squared Error | 13865318318 | 13900048089 | 19028002324 | 16001369118 | 45735927158 | 59478651920 | 13739249915 | 17494080364 | 26923267263 | 15629890357 | 15621776235 | 14269022695 | 13606173540 | 13865318009 |

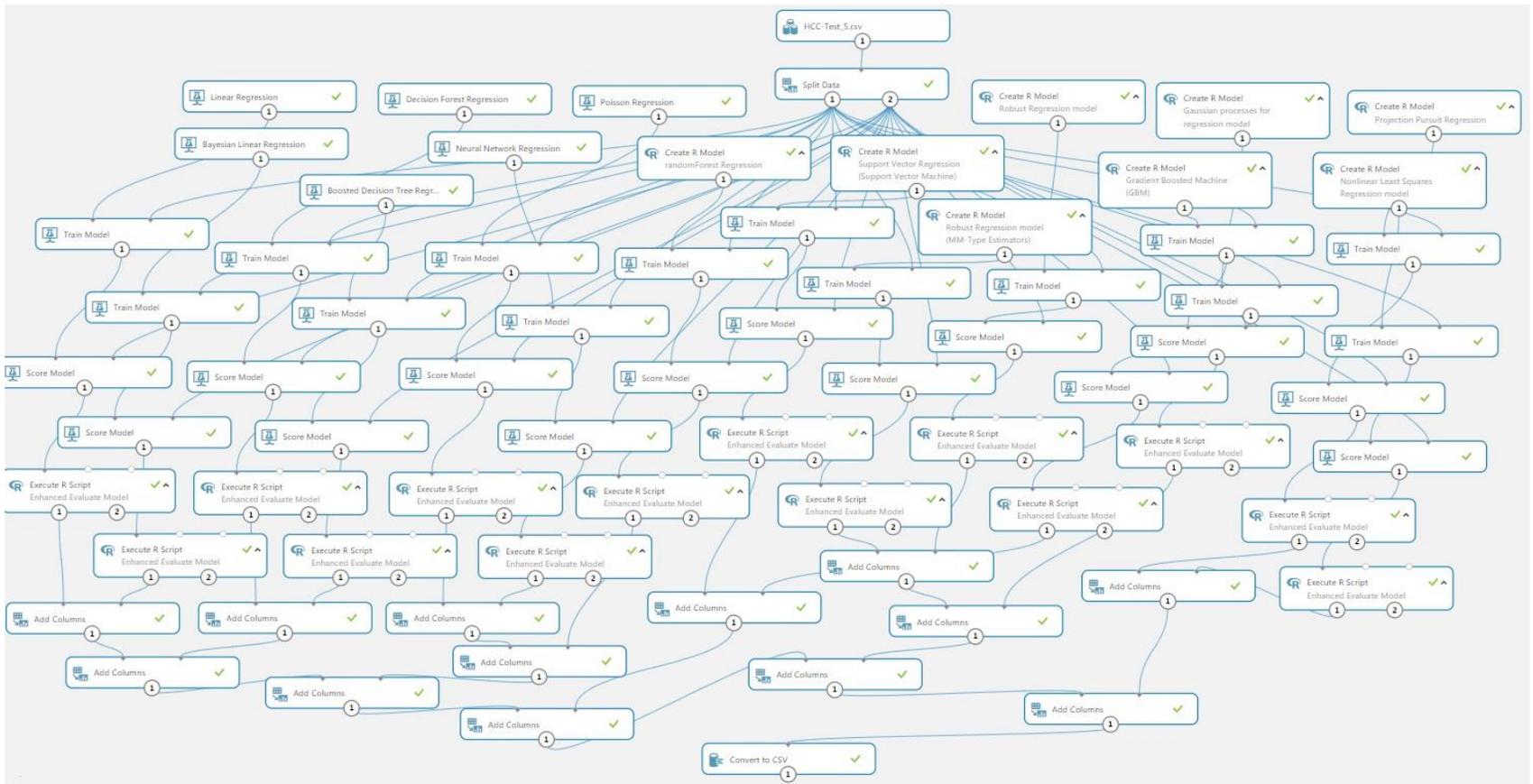

**Fig. 6 Combined workflow of Azure built-in and R-based regression modules**

## Discussion

**Discussion Phase 1**

Selecting most informative features led to intuitively expected results. By all criteria (see Table 1), RIW feature was identified as the most important followed by LOS (with only one exception: second important by Chi Squared was Diagnosis). The scores given to RIW and LOS were significantly higher than those of other features. At the same time, Intervention, Diagnosis and Gender were found the least important by most criteria. CMG and Age Group had mixed standing in the middle of the scale.

**Discussion Phase 2**

Analysis of the Phase 2 testing suggested removal of some models from further consideration. A quick look at the testing results show that neural network regression and Poisson regression models have significantly lower values of all performance metrics compared to other models.

Another four (4) types of models (i.e. linear regression, Bayesian linear regression, decision forest regression and boosted decision tree regression) were compared by three (3) performance metrics: mean absolute error (MAE), root mean squared error (RMSE) and coefficient of determination(CoD). The results are presented in Table A2.2 in Appendix 2.

Overall, performance metrics of all models display similar association with number of features of the testing data set. Performance improves (i.e. MAE and RMSE decrease and CoD increases) when certain data set features are removed (going from Test 1 to Test 6). Significant improvement is observed going from Test 1 to Test 3 (i.e. diagnosis, intervention, case mix group columns are removed one by one). For the tests from 4 to 6, performance of all models stabilizes at the same ranges: 400-500 dollars for MAE, 1,990-2,300 dollars for RMSE, and 0.65-0.75 for CoD. The only exception is the performance of the DFR model, which has certain drop of results in Test 5. Based on the Phase 2 experiments further testing was conducted with testing data sets 4, 5 and 6. Comparison of individual performance of the models gives mixed results. The best MAE is shown by Decision Forest Regression (DFR). The best RMSE is demonstrated by Linear Regression (LR) and Bayesian Linear Regression (BLR). The best CoD is recorded for Decision Forest Regression (DFR) and Boosted Decision Tree Regression (BDT). To re-iterate, the differences in performance metrics for all models were not significant, and all of them were admitted to Phase 3 experiment with data sets Test 4 to Test 6.

As it was noted, Test 4 to Test 6 contain features with most predictive power, identified in the feature selection exercise. So, the results of the feature selection process and results of the errors analysis in the experiment support each other and confirm that using less informative features is not only leading to inefficiency, but also may directly decrease the performance of prediction.

An attempt was made to combine less informative features (Intervention, Diagnosis, CMG, Gender) into a single column using Azure Principal Component Analysis (PCA) module (located in Scale and Reduce, Data Transform). Combined column (a set of positive and negative numbers) has been added to the data set in Test 4. The experiment did not reveal any improvement of performance by any type of regression. Analysis of this combined column using Filter Based Feature Selection module also shows that the predictive power of the combined feature was negligible (but not zero).

**Discussion Phase 3**

Parameter tuning experiments have shown that Decision Forest Regression and Boosted Tree Regression models are responsive to parameter sweeping. Optimization by the minimum mean average error demonstrated improvement by 5% for Decision Forest Regression and 16.9% for Boosted Tree Regression (see Tables 4 and 5). All other performance metrics improved for optimized models as well. Decrease of the root mean squared error (RMSE) provides evidence that larger errors were less proliferant in the tuned models as large errors contribute the most to the value of the RMSE. Both optimized Decision Forest Regression and Boosted Tree Regression models demonstrate similar absolute values of performance metrics: 402-403 dollars for mean average error and 0.75-0.76% for coefficient of determination.

Parameter tuning experiments with all other types of regression, i.e. linear regression, Bayesian linear regression, neural network regression and Poisson regression, either were not possible or did not produce improvements in performance metrics compared to default options.

**Discussion Phase 4**

Seed number change experiment has shown that algorithms Decision Forest Regression and Boosted Decision Tree Regression are tolerant to the change (Appendix 3). The change in MAE was under 1%.

Changing split ratio from 0.5 to 0.75, i.e. making testing data set larger, led to enhanced performance of most algorithms by MAE (Appendix 4).

Changing number of data set features from initial cost, LOS, RIW (HCC-Test_5) and age group to cost, LOS, RIW (HCC-Test_6), i.e. removing age group column, enhanced MAE of most algorithms by 2-8%, except Bayesian Linear Regression and Random Forest Regression (Appendix 5).

**Discussion Phase 5**

Results of the experiments with all 14 models show that performance metrics, especially MAE, vary in a wide range (see Table 6). In terms of MAE, minimum error is around $250 and maximum is around $550. Robust regression demonstrated the best performance (MAE $255). This result is significantly better than for all other R-based models. Also, robust regression outperforms all Azure MLS built-in models. To confirm this finding, robust regression has been calculated with two different algorithms: *rlm* function from {MASS}, and *lmrob* function from {robustbase}. Both algorithms returned practically same results.

For majority of the methods, coefficient of determination is close to 0.70 - 0.75.

Combined performance metrics for Azure built-in and R-based regression models (considering parameter tuning), show that Robust Regression model performed the best (MAE $255) followed by Boosted Decision Tree Regression ($402) and Decision Forest Regression ($403). Although, the latter two display this performance only with tuning (optimization). These algorithms also displayed good tolerance to changes.

Application of the tool is not limited to cost prediction. The tool can also be used for similar types of predictions, e.g. hospital length of stay prediction, or in sectors other than health care.

**Limitations**

Note that the tool has been built and tested using numerical only data with no n.a. (missing) elements. Certain regression models do not except categorical data and conversion to numerical format may be required.

Note that the data set used in the experiment is **simulated** and no warranties are provided as to the validity of the data and how closely it simulates real-world information.

Note that, because of the previous point, the numerical results of the experiment cannot be used to make actual predictions.

## Concluding Remarks

The practical contribution of this study is in delivering an Azure Machine Learning Studio tool for rapid assessment of multiple types of regression models. The operational tool has been published to the web at Azure MLS (Botchkarev, 2018a). It can be used by researchers and practitioners to reproduce the results of this study, conduct experiments with their own data sets or add more regression models to the framework.

The tool offers environment for comparing 14 types of regression models (linear regression, Bayesian linear regression, decision forest regression, boosted decision tree regression, neural network regression, Poisson regression, Gaussian processes for regression, gradient boosted machine, nonlinear least squares regression, projection pursuit regression, random forest regression, robust regression, robust regression with mm-type estimators, support vector regression) in a unified experiment and presents assessment results in a single table using 22 performance metrics: Coefficient of Determination (CoD), Geometric Mean Relative Absolute Error (GMRAE), Mean Absolute Error (MAE), Mean Absolute Percentage Error (MAPE), Mean Absolute Scaled Error (MASE), Median Absolute Error (MdAE), Median Absolute Percentage Error (MdAPE), Median Relative Absolute Error (MdRAE), Mean Error (ME), Mean Percentage Error (MPE), Mean Relative Absolute Error (MRAE), Mean Squared Error (MSE), Normalized Root Mean Squared Error normalized to the difference between maximum and minimum actual data (NRMSE_mm),  Normalized Root Mean Squared Error normalized to the standard deviation of the actual data (NRMSE_sd), Relative Absolute Error (RAE), Root Median Square Percentage Error (RMdSPE), Root Mean Squared Error (RMSE), Root Mean Square Percentage Error (RMSPE), Relative Squared Error (RSE), Symmetric Mean Absolute Percentage Error (sMAPE ), Symmetric Median Absolute Percentage Error (SMdAPE), Sum of Squared Error (SSE).

Using the developed rapid assessment Azure MLS tool, various types of regression machine learning models were evaluated for performing hospital case cost prediction. The higher performance was demonstrated by robust regression model, and tuned versions of the boosted decision tree regression and decision forest regression.

**Acknowledgement**

# References

Alpenberg, J., & Scarbrough, D. P. (2015). Lean Healthcare and Ontario Case Costing - An Examination of Strategic Change and Management Control Systems (December 14, 2014). 2015 *Canadian Academic Accounting Association (CAAA) Annual Conference*. Available at SSRN: *https://ssrn.com/abstract=2538288* or *http://dx.doi.org/10.2139/ssrn.2538288*

Botchkarev, A., & Andru, P. (2008, May). Using financial modelling for integrated healthcare databases. In *Electrical and Computer Engineering, 2008. CCECE 2008. Canadian Conference on* (pp. 001973-001976). IEEE.

Botchkarev, A. (2018a). Revision 2 Integrated tool for rapid assessment of multi-type regression machine learning models. Experiment in Microsoft Azure Machine Learning Studio. Azure AI Gallery. *https://gallery.azure.ai/Experiment/Revision-2-Integrated-tool-for-rapid-assessment-of-multi-type-regression-machine-learning-models*

Botchkarev, A. (2018b). Enhanced model evaluation with multiple performance metrics for regression analysis. Experiment in Microsoft Azure Machine Learning Studio. Azure AI Gallery. *https://gallery.azure.ai/Experiment/Enhanced-model-evaluation-with-multiple-performance-metrics-for-regression-analysis*

Botchkarev, A. (2018c). Evaluating Performance of Regression Machine Learning Models Using Multiple Error Metrics in Azure Machine Learning Studio (May 12, 2018). Available at SSRN: *http://ssrn.com/abstract=3177507*

caret: Classification and Regression Training. R package. *https://cran.r-project.org/web/packages/caret/index.html*

Corral, M., Ferko, N., Hogan, A., Hollmann, S. S., Gangoli, G., Jamous, N., ... & Kocharian, R. (2016). A hospital cost analysis of a fibrin sealant patch in soft tissue and hepatic surgical bleeding. *ClinicoEconomics and outcomes research: CEOR*, *8*, 507.

e1071: Misc Functions of the Department of Statistics, Probability Theory Group (Formerly: E1071). R package. *https://cran.r-project.org/web/packages/e1071/index.html*

kernlab: Kernel-Based Machine Learning Lab. R package. *https://cran.r-project.org/web/packages/kernlab/index.html*

MASS: Support Functions and Datasets for Venables and Ripley's MASS. R package. *https://cran.r-project.org/web/packages/MASS/index.html*

MSBVAR: Markov-Switching, Bayesian, Vector Autoregression Models. R package. *https://cran.r-project.org/web/packages/MSBVAR/index.html*

Osnabrugge, R. L., Speir, A. M., Head, S. J., Jones, P. G., Ailawadi, G., Fonner, C. E., ... & Rich, J. B. (2014). Prediction of costs and length of stay in coronary artery bypass grafting. *The Annals of thoracic surgery*, *98*(4), 1286-1293. DOI: *https://doi.org/10.1016/j.athoracsur.2014.05.073*

randomForest: Breiman and Cutler's Random Forests for Classification and Regression. R package. *https://cran.r-project.org/web/packages/randomForest/index.html*

robustbase: Basic Robust Statistics. R package. *https://cran.r-project.org/web/packages/robustbase/index.html*

Srinivasan, K., Currim, F., & Ram, S. (2017). Predicting High Cost Patients at Point of Admission using Network Science. *IEEE Journal of Biomedical and Health Informatics*, volume: PP, issue: 99. DOI: 10.1109/JBHI.2017.2783049

Sushmita, S., Newman, S., Marquardt, J., Ram, P., Prasad, V., Cock, M. D., & Teredesai, A. (2015, May). Population cost prediction on public healthcare datasets. In *Proceedings of the 5th International Conference on Digital Health 2015* (pp. 87-94). ACM.

stats: The R Stats Package. R package. *https://stat.ethz.ch/R-manual/R-devel/library/stats/html/00Index.html*

Teague, L., Mahoney, J., Goodman, L., Paulden, M., Poss, J., Li, J., ... & Krahn, M. (2011). Support surfaces for intraoperative prevention of pressure ulcers in patients undergoing surgery: a cost-effectiveness analysis. *Surgery*, *150*(1), 122-132.

# Appendix 1. Data Set Characteristics

**Table A1.1**

| Intervention | Diagnosis | Case Mix Group (CMG) | Gender | Age Group | Relative Intensity Weight (RIW) | Length of Stay (LOS) | Cost |
|---|---|---|---|---|---|---|---|
| 1GJ50BA | C3480 | 129 | F | 11 | 1.463 | 3 | 1,480 |
| 1GJ50BA | C781 | 129 | M | 11 | 1.463 | 3 | 1,510 |
| 1FU89NZ | C783 | 129 | F | 15 | 1.405 | 2 | 1,460 |
| 3GY20WC | A157 | 135 | F | 6 | 1.499 | 19 | 3,200 |
| 3GY20WA | A150 | 135 | M | 17 | 2.135 | 13 | 2,230 |
| 3AN20WA | A152 | 135 | M | 9 | 1.499 | 4 | 1,540 |

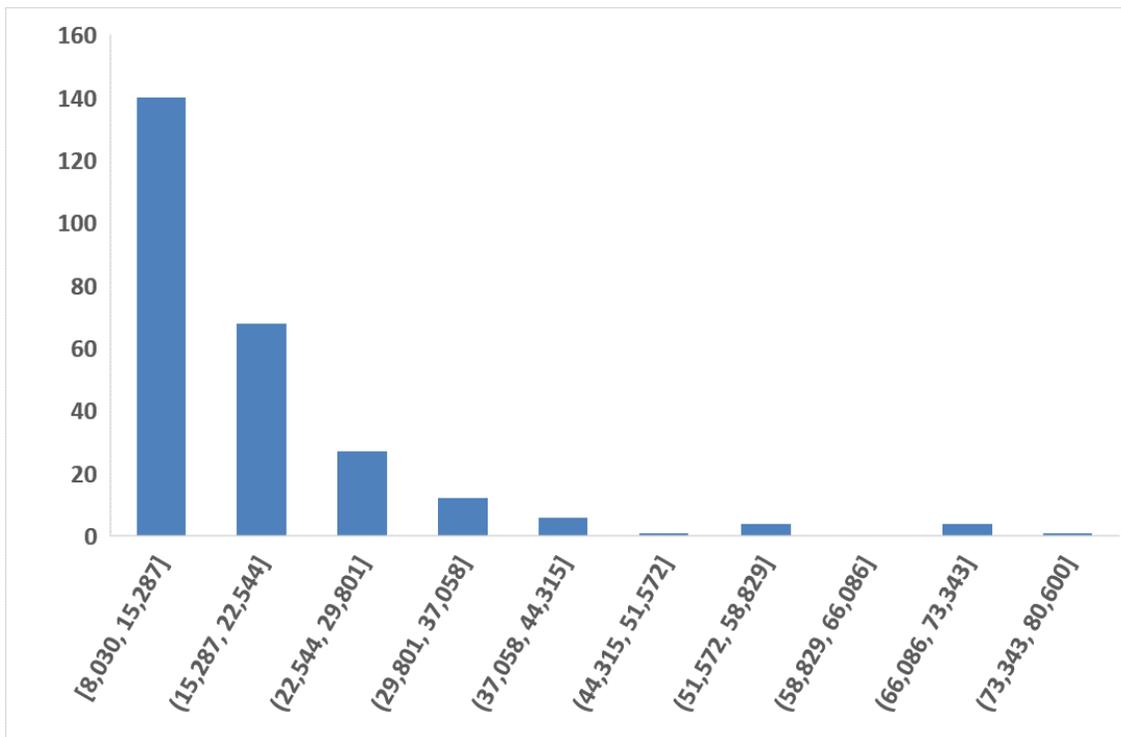

**Fig. A1.1 Histogram of large costs**

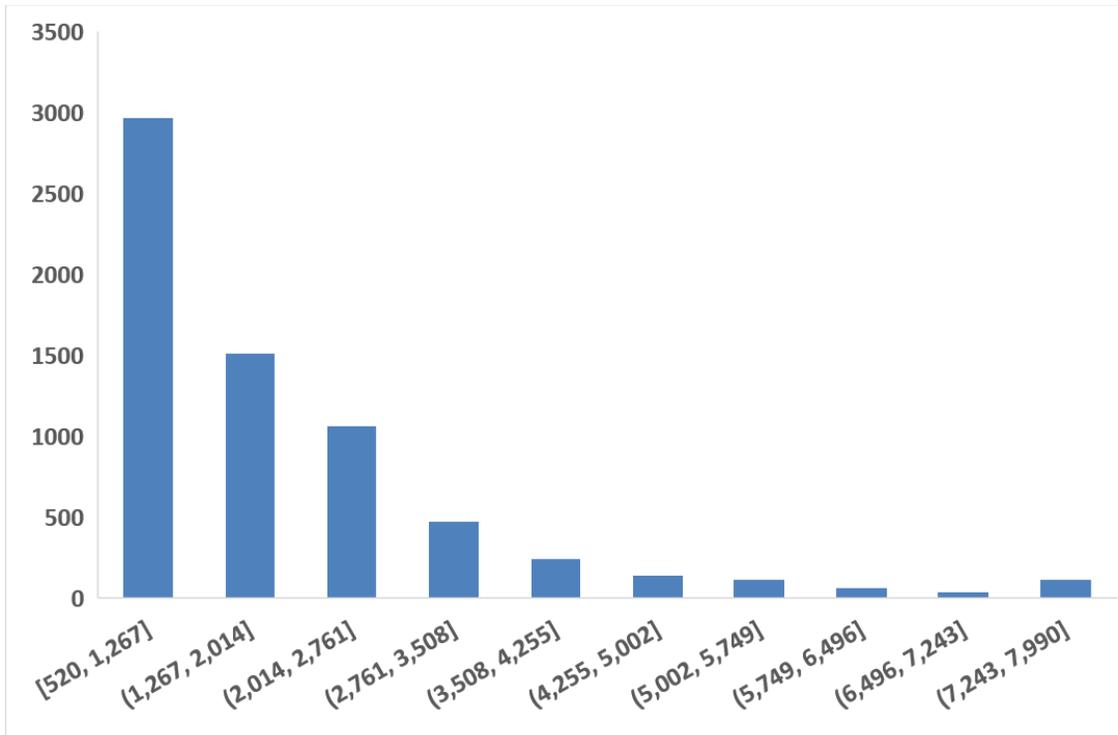

**Fig. A1.2 Histogram of lower costs**

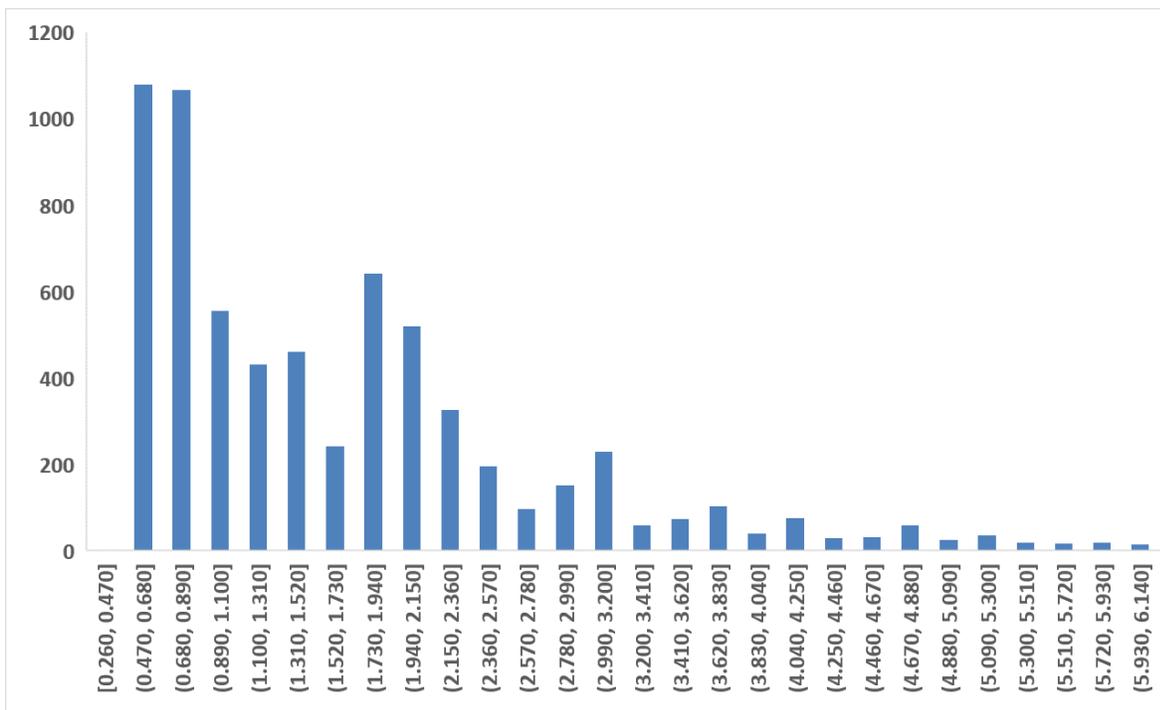

**Fig. A1.3 Histogram of Relative Intensity Weights**

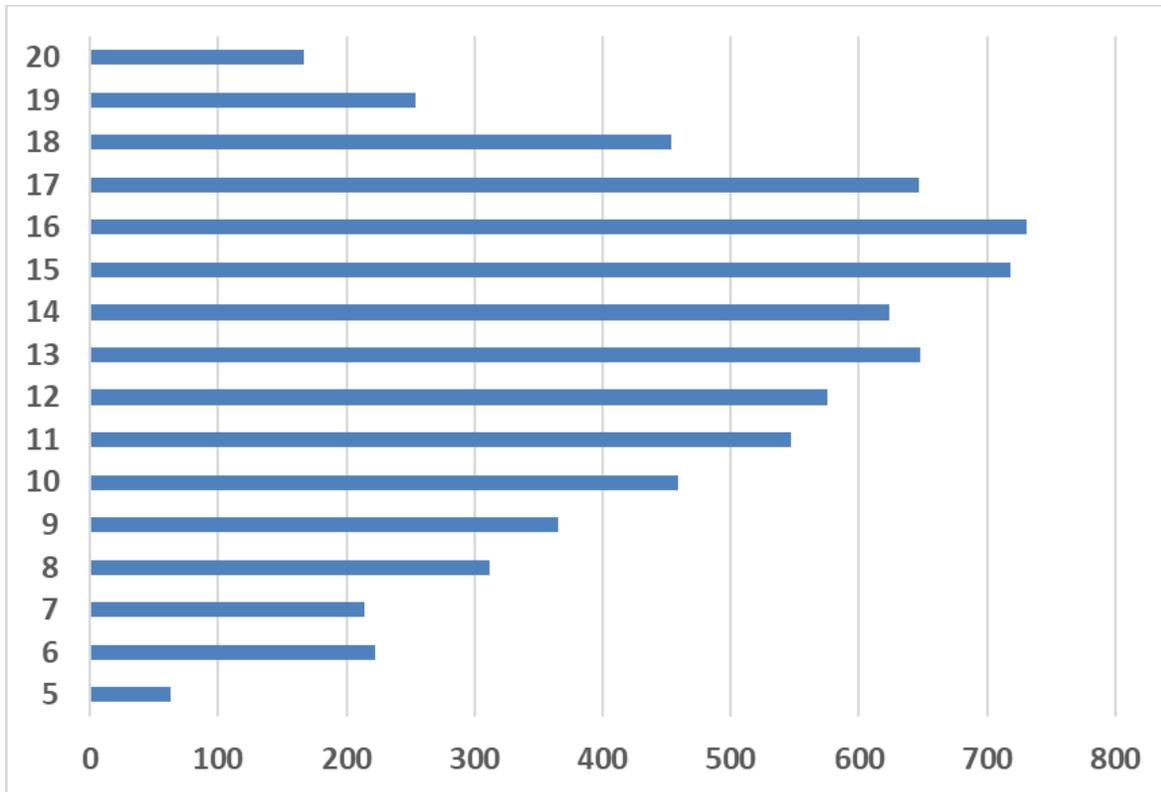

**Fig. A1.4 Distribution of Age Groups**

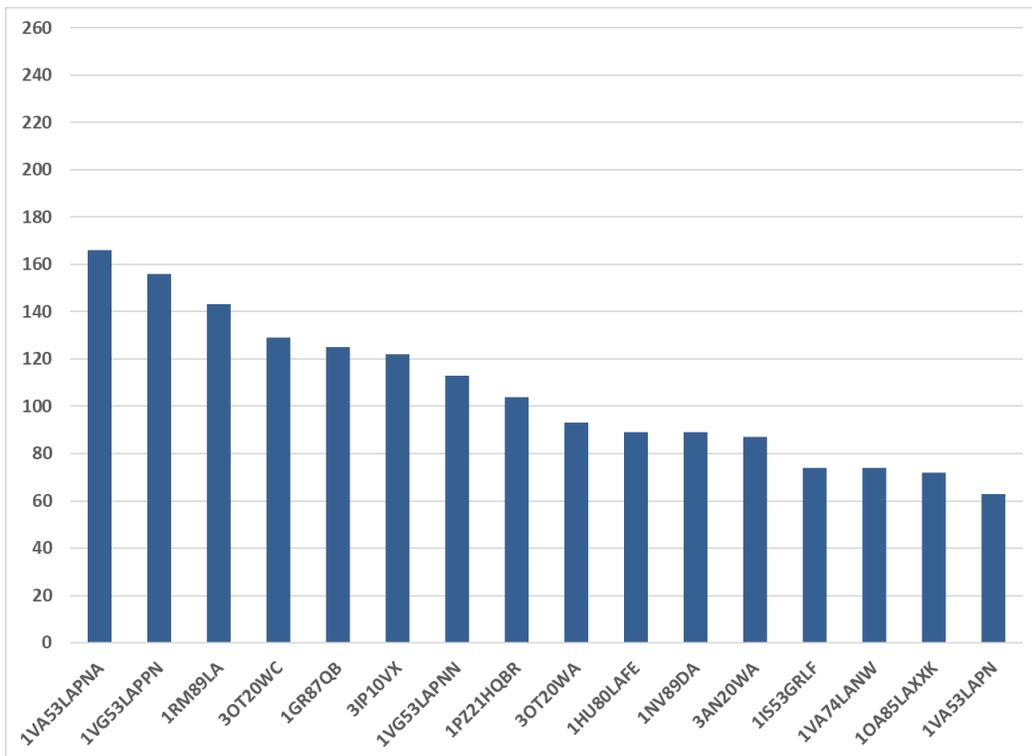

**Fig. A1.5 Distribution of Interventions (60 or more cases)**

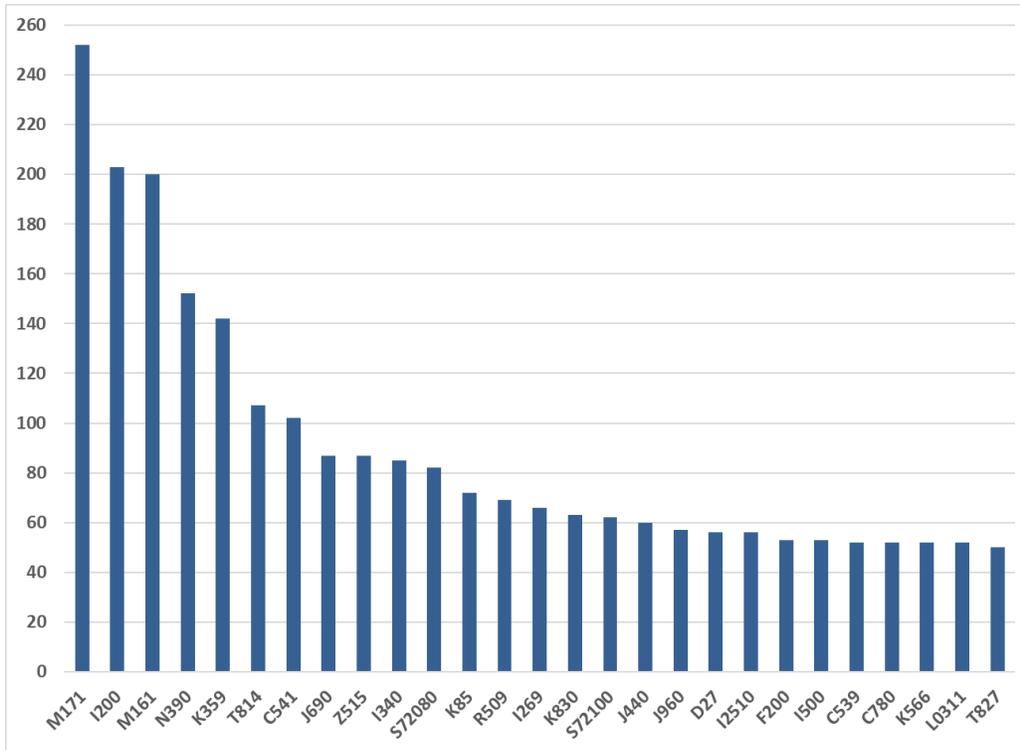

**Fig. A1.6 Distribution of Diagnoses (50 or more cases)**

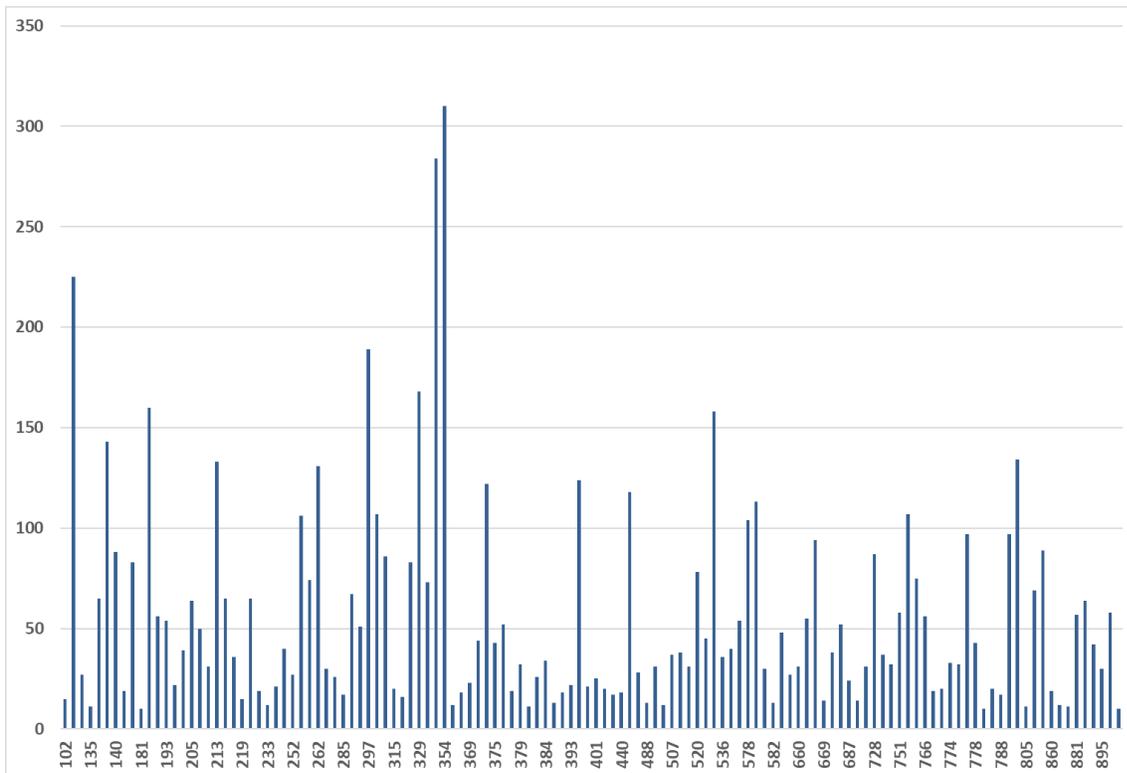

**Fig. A1.7 Distribution of Numbers of Cases by CMG (50 or more cases)**

# Appendix 2. Phase 2 Experiment
## Table A2.1

**Linear**

|  | Test_1 | Test_2 | Test_3 | Test_4 | Test_5 | Test_6 |
|---|---|---|---|---|---|---|
| Mean Absolute Error | 1,021.63 | 670.6467 | 482.412255 | 482.443797 | 481.747978 | 471.563505 |
| Root Mean Squared Error | 2,967.65 | 2,198.52 | 1991.121625 | 1991.249278 | 1990.356631 | 1990.329583 |
| Relative Absolute Error | 0.4716 | 0.3607 | 0.259489 | 0.259506 | 0.259131 | 0.253653 |
| Relative Squared Error | 0.4464 | 0.3045 | 0.24972 | 0.249752 | 0.249529 | 0.249522 |
| Coefficient of Determination | 0.5536 | 0.6955 | 0.75028 | 0.750248 | 0.750471 | 0.750478 |

**Bayesian Linear**

|  | Test_1 | Test_2 | Test_3 | Test_4 | Test_5 | Test_6 |
|---|---|---|---|---|---|---|
| Mean Absolute Error | 739.8899 | 576.1372 | 482.28393 | 482.311151 | 473.495256 | 476.96975 |
| Root Mean Squared Error | 2,364.67 | 2,044.82 | 1991.130386 | 1991.273283 | 1992.847789 | 1994.770778 |
| Relative Absolute Error | 0.3416 | 0.3099 | 0.25942 | 0.259434 | 0.254692 | 0.256561 |
| Relative Squared Error | 0.2834 | 0.2634 | 0.249723 | 0.249758 | 0.250154 | 0.250637 |
| Coefficient of Determination | 0.7166 | 0.7366 | 0.750277 | 0.750242 | 0.749846 | 0.749363 |
| Negative Log Likelihood | 24,916.53 | 32,106.44 | 32321.59 | 32323.42614 | 32331.75165 | 32342.06309 |

**Neural Network**

|  | Test_1 | Test_2 | Test_3 | Test_4 | Test_5 | Test_6 |
|---|---|---|---|---|---|---|
| Mean Absolute Error | 2,532.85 | 1,594.87 | 1601.497564 | 1748.52086 | 1594.829661 | 1607.3112 |
| Root Mean Squared Error | 4,483.38 | 4,094.28 | 4063.073195 | 4325.924503 | 4122.365892 | 4162.185279 |
| Relative Absolute Error | 1.1693 | 0.8579 | 0.861443 | 0.940527 | 0.857856 | 0.86457 |
| Relative Squared Error | 1.0189 | 1.0559 | 1.039843 | 1.178735 | 1.070413 | 1.091192 |
| Coefficient of Determination | -0.0189 | -0.0559 | -0.039843 | -0.178735 | -0.070413 | -0.091192 |

**Decision Forest**

|  | Test_1 | Test_2 | Test_3 | Test_4 | Test_5 | Test_6 |
|---|---|---|---|---|---|---|
| Mean Absolute Error | 644.997 | 597.7595 | 440.321772 | 401.660865 | 424.222292 | 389.802885 |
| Root Mean Squared Error | 2,598.89 | 2,362.32 | 2155.820307 | 2117.165828 | 2331.645791 | 2081.307669 |
| Relative Absolute Error | 0.3469 | 0.3215 | 0.236848 | 0.216053 | 0.228189 | 0.209674 |
| Relative Squared Error | 0.4254 | 0.3515 | 0.292741 | 0.282337 | 0.342439 | 0.272854 |
| Coefficient of Determination | 0.5746 | 0.6485 | 0.707259 | 0.717663 | 0.657561 | 0.727146 |
| Negative Log Likelihood | 27,291.69 | 26,583.40 | 17,439,424,124.15 | 15,080,721,829.81 | 112,070,075,644.37 | 15,426,891,145.67 |

**Poisson**

|  | Test_1 | Test_2 | Test_3 | Test_4 | Test_5 | Test_6 |
|---|---|---|---|---|---|---|
| Mean Absolute Error | 1,772.26 | 1,601.68 | 1617.676485 | 1618.231891 | 1620.586192 | 1635.02305 |
| Root Mean Squared Error | 3,883.55 | 3,586.31 | 3617.879354 | 3617.419128 | 3614.886964 | 3628.874961 |
| Relative Absolute Error | 0.8181 | 0.8615 | 0.870146 | 0.870444 | 0.871711 | 0.879476 |
| Relative Squared Error | 0.7645 | 0.8101 | 0.824454 | 0.824245 | 0.823091 | 0.829473 |
| Coefficient of Determination | 0.2355 | 0.1899 | 0.175546 | 0.175755 | 0.176909 | 0.170527 |

**Boosted Decision Tree**

|  | Test_1 | Test_2 | Test_3 | Test_4 | Test_5 | Test_6 |
|---|---|---|---|---|---|---|
| Mean Absolute Error | 650.26 | 491.41 | 505.035206 | 486.052036 | 484.305374 | 451.464217 |
| Root Mean Squared Error | 2,576.34 | 2,153.41 | 2207.492363 | 2142.714129 | 2138.181411 | 2136.901733 |
| Relative Absolute Error | 0.300186 | 0.264327 | 0.271658 | 0.261447 | 0.260507 | 0.242842 |
| Relative Squared Error | 0.336461 | 0.292086 | 0.306942 | 0.289192 | 0.28797 | 0.287626 |
| Coefficient of Determination | 0.663539 | 0.707914 | 0.693058 | 0.710808 | 0.71203 | 0.712374 |

|  | Test_1 | Test_2 | Test_3 | Test_4 | Test_5 | Test_6 |
|---|---|---|---|---|---|---|
| Features used in tests | Cost | Cost | Cost | Cost | Cost | Cost |
|  | RIW | RIW | RIW | RIW | RIW | RIW |
|  | LOS | LOS | LOS | LOS | LOS | LOS |
|  | Age_gr | Age_gr | Age_gr | Age_gr | Age_gr |  |
|  | Gender | Gender | Gender | Gender |  |  |
|  | CMG | CMG | CMG |  |  |  |
|  | Diagn | Diagn |  |  |  |  |
|  | Interv |  |  |  |  |  |

## Table A2.2

**Mean Absolute Error**

|  | Test_1 | Test_2 | Test_3 | Test_4 | Test_5 | Test_6 |
|---|---|---|---|---|---|---|
| Linear (LR) | 1022 | 671 | 482 | 482 | 482 | 472 |
| Bayesian Linear (BLR) | 740 | 576 | 482 | 482 | 473 | 477 |
| Decision Forest (DFR) | 645 | 598 | 440 | 402 | 424 | 390 |
| Boosted Decision Tree (BDT) | 650 | 491 | 505 | 486 | 484 | 451 |

**Root Mean Squared Error**

|  | Test_1 | Test_2 | Test_3 | Test_4 | Test_5 | Test_6 |
|---|---|---|---|---|---|---|
| Linear (LR) | 2,968 | 2,199 | 1,991 | 1,991 | 1,990 | 1,990 |
| Bayesian Linear (BLR) | 2,365 | 2,045 | 1,991 | 1,991 | 1,993 | 1,995 |
| Decision Forest (DFR) | 2,599 | 2,362 | 2,156 | 2,117 | 2,332 | 2,081 |
| Boosted Decision Tree (BDT) | 2,576 | 2,153 | 2,207 | 2,143 | 2,138 | 2,137 |

**Coefficient of Determination**

|  | Test_1 | Test_2 | Test_3 | Test_4 | Test_5 | Test_6 |
|---|---|---|---|---|---|---|
| Linear (LR) | 0.55 | 0.70 | 0.75 | 0.75 | 0.75 | 0.75 |
| Bayesian Linear (BLR) | 0.72 | 0.74 | 0.75 | 0.75 | 0.75 | 0.75 |
| Decision Forest (DFR) | 0.57 | 0.65 | 0.71 | 0.72 | 0.66 | 0.73 |
| Boosted Decision Tree (BDT) | 0.66 | 0.71 | 0.69 | 0.71 | 0.71 | 0.71 |
| Features used in tests | Cost<br>RIW<br>LOS<br>Age_gr<br>Gender<br>CMG<br>Diagn<br>Interv | Cost<br>RIW<br>LOS<br>Age_gr<br>Gender<br>CMG<br>Diagn | Cost<br>RIW<br>LOS<br>Age_gr<br>Gender<br>CMG | Cost<br>RIW<br>LOS<br>Age_gr<br>Gender | Cost<br>RIW<br>LOS<br>Age_gr | Cost<br>RIW<br>LOS |

**Appendix 3. Change of performance metrics values due to seed number change***

|  | Absolute Change |  |  |  |  | Relative Change |  |  |  |  |
|---|---|---|---|---|---|---|---|---|---|---|
| Algorithm | Mean Absolute Error | Root Mean Squared Error | Relative Absolute Error | Relative Squared Error | Coefficient of Determination | Mean Absolute Error | Root Mean Squared Error | Relative Absolute Error | Relative Squared Error | Coefficient of Determination |
| Bayesian Linear Regression | -97 | -86.97 | -0.06 | 0.00 | 0.00 | -20.55% | -4.36% | -22.64% | -1.65% | 0.55% |
| Boosted Decision Tree Regression | 1 | -291.05 | 0.00 | -0.06 | 0.06 | 0.30% | -13.61% | -1.43% | -20.47% | 8.28% |
| Decision Forest Regression | -2 | 58.55 | -0.01 | 0.04 | -0.04 | -0.44% | 2.51% | -2.19% | 11.30% | -5.88% |
| Linear Regression | -84 | -93.42 | -0.05 | -0.01 | 0.01 | -17.45% | -4.69% | -19.50% | -2.30% | 0.76% |
| Neural Network Regression | -27 | -32.98 | -0.03 | 0.06 | -0.06 | -1.71% | -0.80% | -3.48% | 5.17% | 78.60% |
| Poisson Regression | -801 | -58225.07 | -0.45 | -223.99 | 223.99 | -49.44% | -1610.70% | -52.04% | -27212.98% | 126611.77% |

*Seed number was changed in all modules from 12345 to 98642.
Data set HCC-Test_5.
Split 0.5

**Appendix 4. Change of performance metrics values due to change of split ratio***

|  | Absolute Change |  |  |  |  | Relative Change |  |  |  |  |
|---|---|---|---|---|---|---|---|---|---|---|
| Algorithm | Mean Absolute Error | Root Mean Squared Error | Relative Absolute Error | Relative Squared Error | Coefficient of Determination | Mean Absolute Error | Root Mean Squared Error | Relative Absolute Error | Relative Squared Error | Coefficient of Determination |
| Bayesian Linear Regression | 10 | 248 | 0.00 | 0.03 | -0.03 | 2.2% | 12.5% | -0.8% | 10.7% | -3.6% |
| Boosted Decision Tree Regression | 47 | 340 | 0.02 | 0.05 | -0.05 | 9.7% | 15.9% | 7.0% | 17.6% | -7.1% |
| Decision Forest Regression | 13 | 510 | 0.00 | 0.10 | -0.10 | 3.1% | 21.9% | 0.2% | 28.9% | -15.0% |
| Linear Regression | 11 | 249 | 0.00 | 0.03 | -0.03 | 2.3% | 12.5% | -0.7% | 10.8% | -3.6% |
| Neural Network Regression | -70 | 419 | -0.06 | 0.06 | -0.06 | -4.4% | 10.2% | -7.5% | 6.0% | 91.0% |
| Poisson Regression | 71 | 361 | 0.01 | 0.05 | -0.05 | 4.4% | 10.0% | 1.5% | 5.6% | -26.0% |

*Split ratio was changed from 0.5 to 0.75 (increasing testing data set).
Data set HCC-Test_5.
Seed number 12345.

**Appendix 5. Change of performance metrics values due to change of number of data set features** *

| | Absolute Change | | | | | Relative Change | | | | |
|---|---|---|---|---|---|---|---|---|---|---|
| Algorithm | Mean Absolute Error | Root Mean Squared Error | Relative Absolute Error | Relative Squared Error | Coefficient of Determination | Mean Absolute Error | Root Mean Squared Error | Relative Absolute Error | Relative Squared Error | Coefficient of Determination |
| Bayesian Linear Regression | -3 | -2 | 0.00 | 0.00 | 0.00 | -0.7% | -0.1% | -0.7% | -0.2% | 0.1% |
| Boosted Decision Tree Regression | 33 | 1 | 0.02 | 0.00 | 0.00 | 6.8% | 0.1% | 6.8% | 0.1% | 0.0% |
| Decision Forest Regression | 34 | 250 | 0.02 | 0.07 | -0.07 | 8.1% | 10.7% | 8.1% | 20.3% | -10.6% |
| Linear Regression | 10 | 0 | 0.01 | 0.00 | 0.00 | 2.1% | 0.0% | 2.1% | 0.0% | 0.0% |
| Poisson Regression | -14 | -14 | -0.01 | -0.01 | 0.01 | -0.9% | -0.4% | -0.9% | -0.8% | 3.6% |
| Random Forest Regression | 38 | -30 | 0.03 | 0.00 | 0.01 | 8.2% | -1.5% | 11.5% | 0.9% | 1.0% |

* Changing number of data set features from initial cost, LOS, RIW (HCC-Test_5) and age group to cost, LOS, RIW (HCC-Test_6), i.e. removing age group column.
Seed number 12345.
Split ratio 0.75.

**Appendix 6. Abbreviations**

| | |
|---|---|
| Age_gr | Age group |
| BDTR | Boosted Decision Tree Regression |
| BLR | Bayesian Linear Regression |
| CMG | Case Mix Group |
| CoD | Coefficient of Determination |
| DFR | Decision Forest Regression |
| Diagn | Diagnosis |
| GBM | Gradient Boosted Machine |
| GMRAE | Geometric Mean Relative Absolute Error |
| GPR | Gaussian Processes for Regression |
| Interv | Intervention |
| LOS | Length of stay |
| LR | Linear Regression |
| MAE | Mean Absolute Error |
| MAPE | Mean Absolute Percentage Error |
| MASE | Mean Absolute Scaled Error |
| MdAE | Median Absolute Error |
| MdAPE | Median Absolute Percentage Error |
| MdRAE | Median Relative Absolute Error |
| ME | Mean Error |
| MPE | Mean Percentage Error |
| MRAE | Mean Relative Absolute Error |
| MSE | Mean Squared Error |
| NLSR | Nonlinear Least Squares Regression |
| NNR | Neural Network Regression |
| NRMSE_mm | Normalized Root Mean Squared Error (normalized to the difference between maximum and minimum actual data) |
| NRMSE_sd | Normalized Root Mean Squared Error (normalized to the standard deviation of the actual data) |
| PPR | Projection Pursuit Regression |
| PR | Poisson Regression |
| RAE | Relative Absolute Error |
| RFR | Random Forest Regression |
| RIW | Relative intensity weight |
| RMdSPE | Root Median Square Percentage Error |
| RMSE | Root Mean Squared Error |
| RMSPE | Root Mean Square Percentage Error |
| RR | Robust Regression |
| RR mm | Robust Regression with Mm-Type Estimators |
| RSE | Relative Squared Error |
| sMAPE | Symmetric Mean Absolute Percentage Error |
| SMdAPE | Symmetric Median Absolute Percentage Error |
| SSE | Sum of Squared Error |
| SVR | Support Vector Regression |
| TMH | Tune Model Hyperparameters |